\def\year{2019}\relax
\documentclass[letterpaper]{article} 
\usepackage{aaai19}  
\usepackage{times}  
\usepackage{helvet} 
\usepackage{courier}  
\usepackage[hyphens]{url}  
\usepackage{subcaption}
\usepackage{graphicx} 
\usepackage{multirow}
\usepackage{amsmath}
\usepackage{amsfonts}
\usepackage{breqn}
\usepackage{graphicx}  
\title{Bootstrapping Conditional GANs for Video Game Level Generation}
\author{Ruben Rodriguez Torrado,\textsuperscript{\rm 1 3} Ahmed Khalifa, \textsuperscript{\rm 1} Michael Cerny Green, \textsuperscript{\rm 1 3}\\\Large \textbf{Niels Justesen,}\textsuperscript{\rm 2} \textbf{Sebastian Risi,} \textsuperscript{\rm 2 4} \and \textbf{Julian Togelius}\textsuperscript{\rm 1 3 4} \\
\textsuperscript{1}New York University, \textsuperscript{2}IT University of Copenhagen, \textsuperscript{3}OriGen.ai, \textsuperscript{4}modl.ai\\
rubentorrado@origen.ai, ahmed@akhalifa.com, mike.green@nyu.edu,\\ njustesen@gmail.com, sebastian.risi@gmail.com, julian@togelius.com
}
 
\begin{document}

\maketitle

\begin{abstract}
Generative Adversarial Networks (GANs) have shown impressive results for image generation. However, GANs face challenges in generating contents with certain types of constraints, such as game levels. Specifically, it is difficult to generate levels that have aesthetic appeal and are  playable at the same time. Additionally, because training data usually is limited, it is  challenging to generate unique levels with current GANs. In this paper, we propose a new GAN architecture named \emph{Conditional Embedding Self-Attention Generative Adversarial Network} (CESAGAN) and a new bootstrapping training procedure. The CESAGAN is a modification of the self-attention GAN that incorporates an embedding feature vector input to condition the training of the discriminator and generator. This allows the network to model non-local dependency between game objects, and to count objects. Additionally, to reduce the number of levels necessary to train the GAN, we propose a bootstrapping mechanism in which playable generated levels are added to the training set. The results demonstrate that the new approach does not only generate a larger number of levels that are playable but also generates fewer duplicate levels compared to a standard GAN. 
\end{abstract}

\section{Introduction}
Procedural Content Generation (PCG) is a term defining  methods in which content for games or simulations is created using programmatic means. Procedural Content Generation via Machine Learning (PCGML) is a PCG approach in which a machine learning component is trained on existing content to generate new game content~\cite{summerville2018procedural}. Machine learning approaches have been applied to create game content as varied as platform  levels~\cite{summerville2016super}, strategy game maps~\cite{lee2016predicting}, and collectible cards~\cite{summerville2016mystical}. Different types of machine learning algorithms can be used, as long as they learn some aspect of the distribution of the training set in such a way that new content can be sampled from the model. Machine learning algorithms for PCGML include statistical methods such as n-grams~\cite{dahlskog2014linear} and Markov chains~\cite{snodgrass2016controllable} as well as deep learning techniques such recurrent LSTM networks~\cite{summerville2016super} and convolutional networks~\cite{volz2018evolving}.

Research on PCGML is only a few years old, spurred on by the recent  advancements in  machine learning models for creative or generative tasks. For example, neural networks are able to generate music~\cite{eck2002finding}, faces~\cite{karras2017progressive}, or text~\cite{radford2018better} with  impressive results. At first glance, it seems reasonable that the same methods could be used to generate content such as levels, characters, or quests for your favorite game. However, there are some crucial differences and challenges when dealing with game content.

The first difference is data scarcity. When training a machine learning model to create lifelike faces, coherent news stories or harmonious music, training data is abound. It is easy to find thousands of training examples, and deep learning models in particular achieve increasingly better results with more training data. However, for most games, only a limited amount of content exists. Super Mario Bros has a few dozen levels, Mass Effect probably less than a hundred named characters, Skyrim only tens of non-trivial quests, and Grand Theft Auto V a handful of car models and weapon types. Only relatively few games have user-made content of sufficient quantity and quality to make for a good training set, and this content is often not publicly available (e.g.\ the level corpus from Super Mario Maker).

The second difference is that many types of game content (in particular \emph{necessary content}~\cite{togelius2011search}) have functional requirements. A picture of a face where one eye is smudged out is still recognizably as a face, and a sentence can be agrammatical and misspelled but still readable; these types of content do not need to \emph{function}. In contrast, game levels, in which it is impossible to find the key to the exit are simply unplayable, and it does not matter how aesthetically pleasing they are. The same holds true for a ruleset, which does not specify how characters move, or a car where the wheels do not touch the ground. In this respect, it is useful to think of most game content as being more like program code than like images. The problem is that most generative representations are not intrinsically well-suited to create content with functional requirements. Many such functional requirements depend on counting items or non-local relation (e.g. there must be as many keys and doors but they are in different parts of the level) and these are hard to capture with standard network architectures.

One way of addressing the problem of functional requirements is to combine PCGML with a search-based approach, exploring the space learned by a trained model rather than just sampling from it. In particular, \emph{Latent Variable Evolution}, originally invented to produce fingerprints able to bypass authentication schemes~\cite{bontrager2019deepmasterprints}, was successfully used to generate Super Mario Bros level segments with functional properties~\cite{volz2018evolving}. However, it is  desirable to have a model learn the functional requirements, regardless of whether we later choose to generate through random sampling or search.

This paper introduces a new PCGML method that seeks to incorporate functional requirements while at the same time mitigating training data scarcity. 
At the core of this method is a new GAN architecture, the CESAGAN, which incorporates self-attention to capture nonlocal spatial relationships and a conditional input vector.
In order to help the GAN learn functional requirements, we input to the network not only the raw level geometry but also aggregates of level features. Levels are generated through sampling the generator network, and all generated levels are tested for playability. Levels that are playable, and sufficiently different from the levels in the training set, are added to the training set for continued training of the GAN. This way, the training set is bootstrapped from a very small number of levels to a much larger set with commensurably wider coverage, while training the model to only generate playable levels.
 
\section{Related Work}
\label{related_work}

This section discusses a general overview of procedural content generation (PCG), followed descriptions of applications of GAN-based level generation in video games. 

\subsection{Procedural content generation}
PCG refers to the use of computer algorithms to produce content. These techniques have played an important role in video games since the early eighties, with such examples as Rogue (Glenn Wichman, 1980), Elite (David Braben and Ian Bell, 1984), and Beneath the Apple Manor (Don Worth, 1978). PCG is primarily used in games because of its the ability to produce large amounts of content with a negligible memory cost, fitting on a small floppy disk~\cite{Karth2015Elite}. Although scarcity of memory is less of a concern today, PCG is widely used in game design and development such as Spelunky (Derek Yu, 2008), The Binding of Isaac (Edmund McMillen and Florian Himsl, 2011), or No Mans Sky (Hello Games, 2016). 

Designers and developers are using PCG ~\cite{shaker2016procedural} for a variety of different purposes, such as tailoring game contents to the player's taste, reducing the time and cost for designing and developing games, assisting creative content generation, exploring new types of games, or understanding the design space of games. PCG can be used to generate any type of content from textures to game rules. Some types are easier than the others, such as generating vegetation (trees, bushes etc) in games~\cite{IDV2000SpeedTree}. While level generation might seem like a trivial problem, as it has been used since the early days of video games, this is not the case. Most known level generation algorithms are tailored to generating content for a particular game~\cite{dahlskog2012patterns,taylor2011procedural,hunt2007difficulty,ferreira2014search} using a significant amount of game or genre specific knowledge to make sure the generated levels are playable and enjoyable. 

Recently, there has been an increase in the development of generalized algorithms and methods for PCG~\cite{khalifa2016general,khalifa2015puzzle}. Search-based Procedural Content Generation (SB-PCG)~\cite{togelius2011search} methods use search algorithms to generate levels, applying simulations and automated playthroughs to validate the generated content. Procedural Content Generation via Machine Learning (PCG-ML)~\cite{summerville2018procedural} methods use (small)  example sets of levels to train on, after which they generate new levels. This paper proposes a new PCG-ML approach as a solution to the generality problem present in level generation research.

\subsection{Generative Adversarial Networks }

The architecture of a Generative Adversarial Network (GAN)~\cite{goodfellow2014generative} can be understood as an adversarial game between a generator ($G$), which maps a latent random noise vector to a generated  sample, and a discriminator ($D$), which classifies generated samples as real or fake. These adversaries are trained at the same time, striving towards reaching a state where the discriminator maximizes its ability to classify correctly and the generator learns to create new samples that are good enough to be classified as genuine.  GANs became popular in recent years due to their impressive results in tasks such as image generation. However, training GANs is not a trivial procedure: the training process is often unstable, where the generator produces unrealistic samples, or the discriminator is no more accurate than a coin toss. For these reasons, many extensions have been proposed to improve the training process and the quality of the results. For example, \citeauthor{mirza2014conditional}~\shortcite{mirza2014conditional} feed a vector $y$ to train $G$ and $D$ conditioned to generate descriptive tags which are not part of training labels. In addition, \citeauthor{bellemare13arcade}~\shortcite{bellemare13arcade} proposed a new training methodology to grow both the generator and discriminator architecture complexity progressively, reaching a higher-quality on the CELEBA dataset.

More recently, high-quality results have been reported using attention mechanisms in deep learning~\cite{vaswani2017attention}. Attention mechanisms are a very simple idea that identifies the most relevant variables dynamically in more complex deep neural network architecture such as, convolutional neural network (CNN). Recently, \citeauthor{zhang2018self}~\shortcite{zhang2018self}  combined attention mechanism and GANs to generate and discriminate high-resolution details as a function of only spatially local points in lower-resolution feature maps. 

In this paper, we propose a new GANs architecture which combines both ideas, conditional GANs and attention mechanisms, in other words, we combine attention mechanisms with conditional GANs in order to improve the quality and diversity of generated levels. 

\subsection{PCG and GANs}

\citeauthor{volz2018evolving}~\shortcite{volz2018evolving} and \citeauthor{giacomello2018doom}~\shortcite{giacomello2018doom} first introduced the idea of unsupervised learning techniques for PCG. \citeauthor{giacomello2018doom}~\shortcite{giacomello2018doom} trained a GAN to create plain level images for DOOM by combining image and topological features extracted from human-designed content. Though this methodology generates realistic levels from the topological point of view, the playability of the generated levels was not tested. 

\citeauthor{volz2018evolving}~\shortcite{volz2018evolving} combined GANs with latent variable evolution (LVE)~\cite{bontrager2017deepmasterprint} to optimize the input latent vector of a GAN generator to create levels for Mario Bros. Deep Convolutional GANs (DCGANs) were adapted to generate levels, and CMA-ES searched in the space of latent vectors. The results demonstrated that it is often possible to generate  both realistic and playable levels for Mario Bros video game. 
In this paper, we are comparing our method with the approach introduced by \citeauthor{volz2018evolving}~\shortcite{volz2018evolving}.

\subsection{General Video Game AI Framework}
The General Video Game Artificial Intelligence framework~\cite{perez2016general} (GVG-AI) is a framework built to run 2D arcade-like games written in Video Game Description Language (VGDL)~\cite{ebner2013towards}. Originally developed for game-playing competition, GVG-AI has since evolved to be a primary faucet for a variety of research projects and competitions for game generation \cite{khalifa2017rulegen}, level generation \cite{khalifa2016general}, tutorial generation \cite{green2018atdelfi}, and simple game prototyping. The single-player~\cite{perez20162014}, two-player~\cite{gaina2017gvgai}, and learning~\cite{perez2018general} tracks of the competition have resulted in the creation of more than $200$ controllers and well over $120$ games.


\section{Methodology}

GANs have proven very successful in generating images and similar types of content that do not have structural and functional representation. 
However, when generating game levels, simple GAN approaches have several shortcomings, which the method presented in this paper tries to overcome:  

\begin{enumerate}
\item Reducing the amount of information necessary to train the discriminator; most games have just a few human-designed levels available to form a training set.
\item Increasing quality; GANs often generate levels with low quality that are sometimes unplayable.
\item Increasing diversity; the diversity and number of unique generated levels are limited with previous approaches~\cite{volz2018evolving}.
\end{enumerate}

We approach challenge 1 with a bootstrapping technique and challenges 2 and 3 with a new GAN architecture.

\subsection{Conditional Embedding Self-Attention Generative Adversarial Networks (CESAGANs)}

Previous GAN-based models for level generation are built using convolutional layers. 
A convolutional layer is a local operation whose correlation depends on the spatial size of the kernel. For example, in a convolution operation for level generation, it is hard for an output on the top-left position to have any correlation to the output at bottom-right. A deep convolution network with many layers would be required which will increase the large search space. 

This phenomena has become larger for video games level generation where just three tiles/pixels located far away from each other (e.g.\  avatar-door-key) must be correlated to generate a playable level. An intuitive solution to this problem could be reduce the kernels sizes and layers located deeper in the network to be able to capture this relationship later.
However, this approach would increase the  number of layers of the deep neural network significantly and thus make the GAN training more unstable ~\cite{salimans2016improved,kodali2017convergence}.

\begin{figure*}[ht]
\centering
\includegraphics[width=0.8\linewidth]{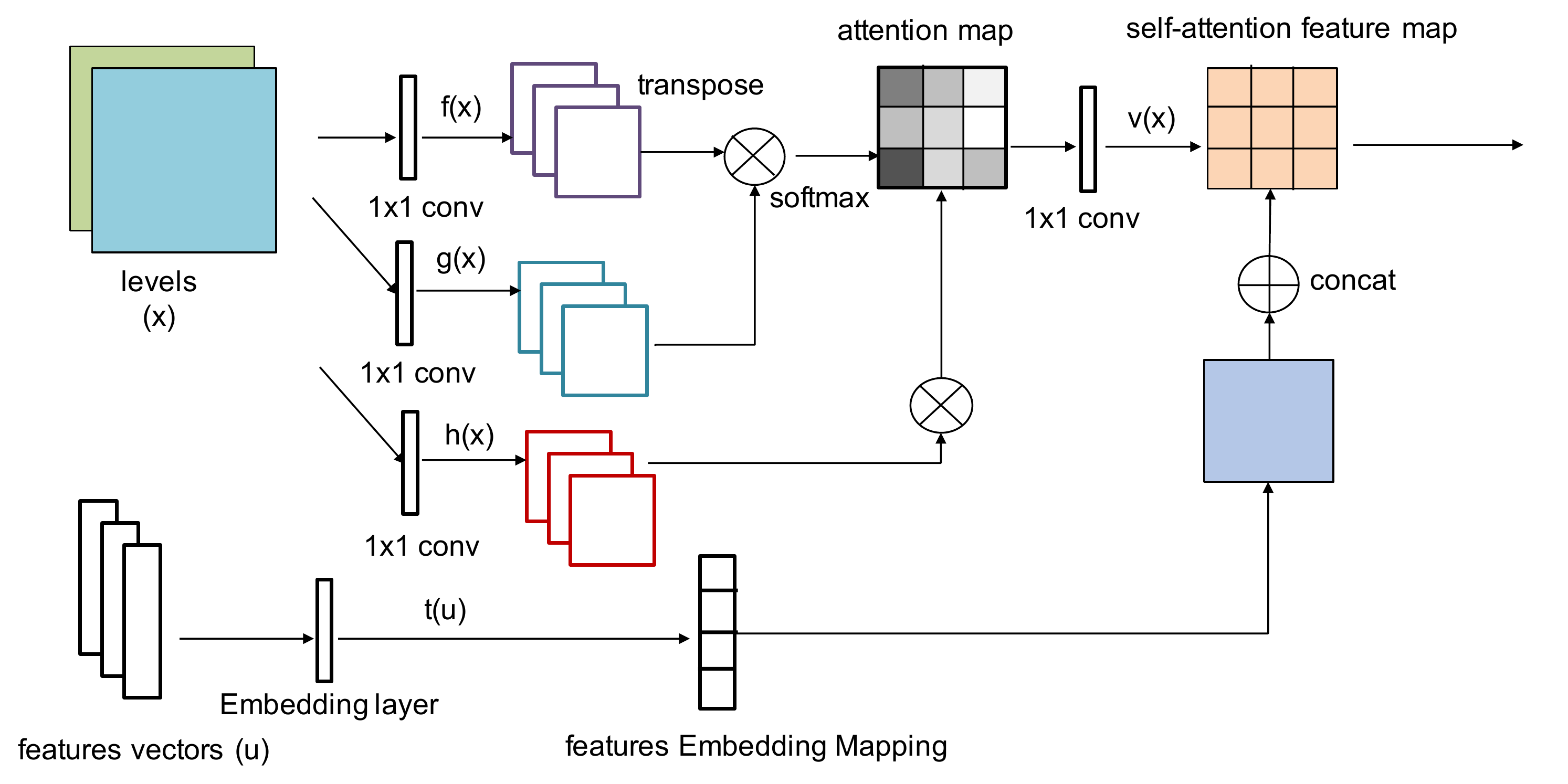}
\caption{\label{fig:architecture} The architecture for our  Conditional Embedding Self-Attention Generative Adversarial Network (CESAGAN). The approach combines a SAGAN \cite{zhang2018self}  architecture (top), with a conditional embedding for the feature information vector $u$ (bottom). We concatenate the feature embedding mapping and the self-attention feature map to combine SAGAN with the  conditional vector representation $u$. This network is applied to both the generator (G) and discriminator (D)} 
\end{figure*}

One potential method that could keep balance between efficiency and capturing long-range dependencies is a self-attention GAN (SAGAN). A Self-attention GAN ~\cite{cheng2016long} is based on three different vectors: query (f), key (g) and value (h) which are three different mappings (e.g.\ output of a single perceptron neural network) of the input data (e.g.\ an image) \ref{fig:architecture}. The query  and key undergo a matrix multiplication then pass through a softmax, which converts the resulting vector into probability distribution attention map.   
This attention map determines the weight of each of the tiles and keep it in memory. Finally, the attention map is multiplied by the value to determine the relationship at a position in a sequence by attending to all positions within the same sequence. It has been shown to be very useful in machine translation ~\cite{vaswani2017attention} and image generation ~\cite{parmar2018image}.

In our experiments, we adapt self-attention GANs ~\cite{zhang2018self} for video game level generation. The mechanism is shown in Figure~\ref{fig:architecture}. The one-hot tile level representation from the hidden layer is transformed into two feature spaces $f$ and $g$ to compute the attention, where $f(x)=W_{f}x$ and  $g(x) = W_{g}x$ are the query and key. We transpose the query and matrix-multiply it by the key, $s_{i,j}=f(x_{i})^{T} g(x_{j})$ and take the softmax on all the rows in order to calculate the attention map:

\begin{equation}
\beta_{j,i}=\frac{exp(s_{i,j})}{\sum_{i=1}^{N}exp(s_{i,j})}
\end{equation}

As we described above, $\beta_{j,i}$ indicates the correlation at a position $i$ when mapping the area $j$. 
Finally, the output of the attention layer is $o_{j}=v(\sum_{i=1}^{N}\beta_{j,i}h(x_{i}))$, 
where $v$ and $h$ are the output of the 1x1 convolutional future. 
This self-attention map layer helps the network capture the fine details from even distant parts of the image and creates a memory for future correlations.  

In SAGAN, the attention module has been used to train the generator and the discriminator, minimizing the hinge version
of the adversarial loss ~\cite{lim2017geometric,zhang2018self}: 
\begin{equation} \label{eq:lossd}
\begin{split}
L_{D} = & -\mathbb{E}_{(x,y)p_{data}}[min(0,-1+D(x,y))]\\
        & -\mathbb{E}_{(z),p_{x}(y),p_{data}}[min(0,-1+D(G(z),y))]
\end{split}
\end{equation}


\begin{equation}\label{eq:lossg}
L_{G}=-\mathbb{E}_{(z),p_{x}(y),p_{data}}D(G(z),y),
\end{equation}
where $z$ is the latent vector. However, this architecture does not guarantee that the generated levels respect different playability-required features such as a minimum/maximum numbers of avatars or enemies. 
For that reason, we extend SAGANs to train the generator and discriminator conditioned to an auxiliary information input feature vector $u$ (e.g.\ class labels or data from other modalities) of each level.  The mapping representation of the feature vector $u$ into the self-attention map is learned by a neural network (the embedding network) during the supervised training process of SAGAN. The embedding network transforms the vector $u$ into a new feature continuous space $t(u)$, where $t(u)=W_{t}u$. 
In the experiments performed in this paper, the feature vector consists of a count of the number of times each sprite appears in the level.
Since our feature vector $u$ is a simple combination of categorical and continuous heuristics, a multilayer perceptron (MLP) was used as the embedding network in this paper. 

Embedding representations reduce memory usage and speed up neural network training when compared with more traditional representations of auxiliary information (feature vector $u$) such as one-hot encoding~\cite{guo2016entity}. In addition, the new representation of $u$ in the embedding space enables the correlation of similar values of categorical variables. Such correlations are more difficult to capture with a simple one-hot representation. This allows the new representation to find more general patterns of the feature vector and therefore allows the SAGAN architecture to generalize better. 

Finally, the output of the embedding mapping $t(u)$ is concatenated with the output of the attention layer $o(i)$, conditioning the adversarial loss functions \ref{eq:lossd} and \ref{eq:lossg} to the input feature vector $u$:

\begin{equation}\label{eq:lossdcond}
\begin{split}
L_{D} & =  -\mathbb{E}_{(x,y),{p}_{data}}[min(0,-1+D((x,y)|\textbf{u}))]\\
      & -\mathbb{E}_{(z),{p}_{x}(y),{p}_{data}}[min(0,-1+D((G(z),y)|\textbf{u}))]
\end{split}
\end{equation}


\begin{equation}\label{eq:lossgcond}
L_{G}=-\mathbb{E}_{(z),p_{x}(y),p_{data}}D((G(z),y)|\textbf{u})
\end{equation}

We call the proposed method Conditional-Embedding Self-Attention Generative Adversarial Networks (CESAGAN) (Figure~\ref{fig:architecture}).  It is important to note that the additional conditioning input of CESAGAN enables the production of levels using specific input information such as number of enemies and player avatars. In other words, the new architecture gives significantly more control to generate game content such as levels with desired characteristics and extends the application areas PCG can be applied to. 

\subsection{Bootstrapping}

Our new architecture CESAGAN has the potential to improve the playability of generated maps. However, we still require a considerable number of training levels for training the discriminator to achieve diversity in the generated levels. For that reason, we have proposed a bootstrapping mechanism to improve the efficiency of CESAGAN architecture explained above. This bootstrapping mechanism takes advantage of built-in game properties that are shared amongst all computer programs, mainly the fact that they can be checked for functionality by attempting to play/execute them. This differentiates game levels from other domains, like pure images.

\begin{figure}[ht]
\centering
\includegraphics[width=1.0\linewidth]{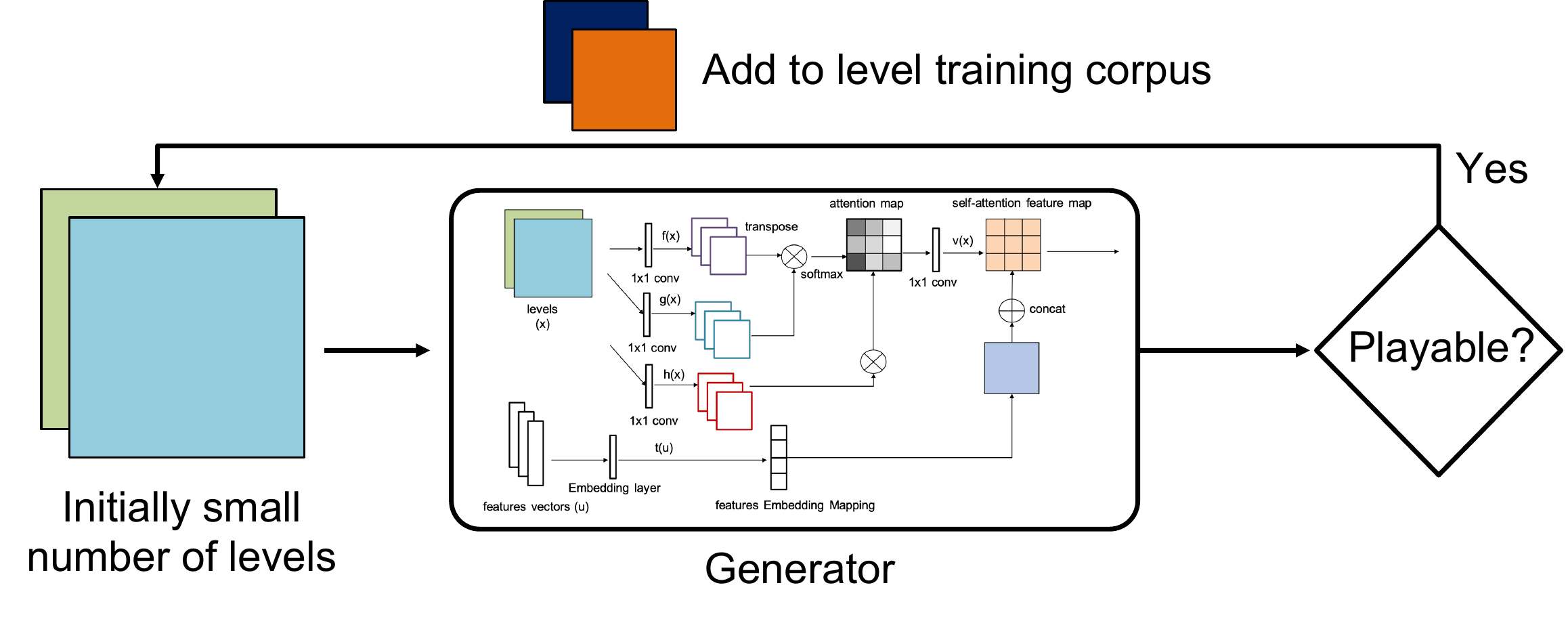}
\caption{\label{fig:architecture2} Conditional Embedding Conditional Self-Attention Generative Adversarial Network (CESAGAN) with bootstrapping. The bootstrapping mechanism increases the number of training examples after passing a playability and diversity test.  Bootstrapping improves the quality of the GAN's discriminator.}
\end{figure}

After each epoch, a new set of levels is generated, on which a playability analysis is carried out to identify unique, playable levels. We propose a set of heuristics for the playability test which is detailed in the Experiments section. Once these levels are identified and scanned for duplicates, the amount of training data $p_{data}$ is increased to $p_{data}^{'}$ for training the generator and the discriminator for the next epoch (see Figure~\ref{fig:architecture2}). Future work could apply AI agents/personas to check for playability of the generated levels in a more dynamic way~\cite{holmgard2018automated}.





\section{Experiments}

The CESAGAN network uses 1$\times$1 convolutions in the discriminator and 1$\times$1 deconvolutions in the generator. Additionally, we employ batchnorm both in the generator and discriminator after each layer and ReLU activations. For the conditional embedding layer, we use a simple fully connected layer that is concatenated with the self-attention feature map. Each type of tile is encoded with an ASCII character in the textual representation of the level and uniquely mapped to a numerical identity. 

\begin{figure}
    \centering
    \includegraphics[width=0.19\linewidth]{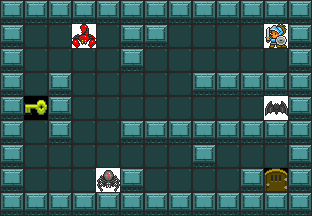}
    \includegraphics[width=0.19\linewidth]{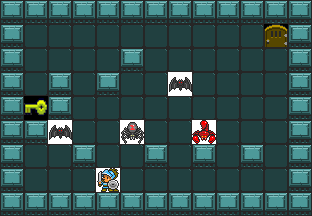}
    \includegraphics[width=0.19\linewidth]{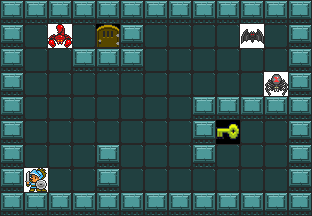}
    \includegraphics[width=0.19\linewidth]{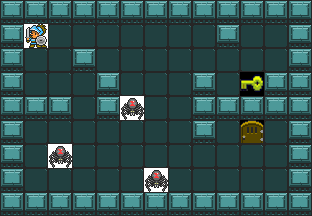}
    \includegraphics[width=0.19\linewidth]{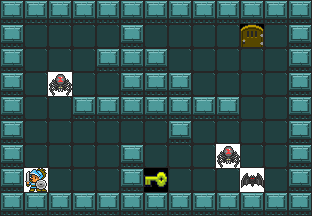}
    \includegraphics[width=0.19\linewidth]{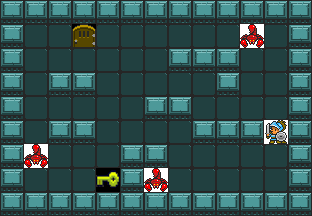}
    \includegraphics[width=0.19\linewidth]{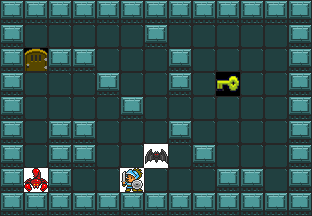}
    \includegraphics[width=0.19\linewidth]{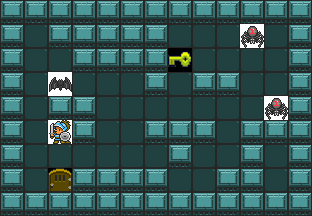}
    \includegraphics[width=0.19\linewidth]{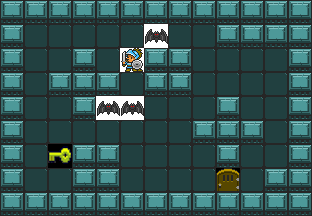}
    \includegraphics[width=0.19\linewidth]{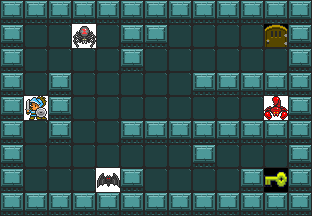}
    \caption{Example of human designed levels used for training. The 5 levels shown at the top are the ones that come with the GVGAI framework.}
    \label{fig:human_levels}
\end{figure}

Finally, both the generator and discriminator are trained with RMSprop with a batch size of 32 and the default learning rate of 0.0001 for 10,000 iterations. To train the discriminator we  used two sets of training levels. The first one consists of 45 human-designed levels, including the five levels that come with the GVGAI framework, while the second one only consists of the five  human-designed levels from the GVGAI framework. Figure~\ref{fig:human_levels} shows some examples of the training data, where the top five images are the levels that come with the GVGAI framework.

\begin{table}[ht]
    \centering
    \begin{tabular}{|l|c|c|}
        \hline
        Tile type & Symbol & Identity \\
        \hline
        Wall & w & 0\\
        Empty & . & 1\\
        Key & + & 2\\
        Exit door & g & 3\\
        Enemy 1 & 1 & 4\\
        Enemy 2 & 2 & 5\\
        Enemy 3 & 3 & 6\\
        Player & A & 7\\
        \hline
    \end{tabular}
    \caption{Mapping of Zelda encoding tiles. Each symbol is encoded as a on-hot encoding for the GAN.}
    \label{tab:tiles}
\end{table}

To evaluate the presented approach we use the game Zelda from the GVGAI environment~\cite{gaina2017gvgai}. This game is a VGDL port of the dungeon system from ``The Legend of Zelda'' (Nintendo, 1986). In this game, the player needs to collect a key and reach the exit door without getting killed by the moving enemies. The player can kill the enemies using their sword for extra points. Table~\ref{tab:tiles} shows the encoding for the tiles in Zelda. Our baseline is the adaptation of GAN architecture proposed by \citeauthor{volz2018evolving}~\shortcite{volz2018evolving} for Zelda. In order to compare both approaches we generated 15,000 levels for both models.

For the bootstrapping playability check, the following seven heuristics are used; they are based on our knowledge about the game to ensure playability (Figure~\ref{fig:human_levels}): 
\begin{itemize}
    \item There is only one player avatar.
    \item There is only one key.
    \item There is only one door.
    \item Enemies cover less than 60\% of the empty space (it is harder to beat the level when there are too many enemies)
    \item The avatar can reach the key using an A* algorithm.
    \item The avatar can reach the door using an A* algorithm.
    \item The level has a border of walls to prevent the avatar and enemies to go outside the level.
\end{itemize}
For more generality, one could use automated game playing agents. For example, our system could have used a planning agent from the GVGAI framework~\cite{perez2018general} to check for playability, rather than heuristics. However, doing so would have increased the amount of time that it takes to train, as the agents have to play each level.

\section{Results}

In order to evaluate the efficiency of the GAN models, generated levels are tested for playability and duplication for CESAGAN with and without bootstrapping. They are compared with the state-of-art techniques below. We define two sets of training data: (1) five human-generated levels and (2) 45 human-generated levels. 

\begin{table}
  \centering
  \setlength\tabcolsep{2.2pt}
  \begin{tabular}{|l|c|c|c|}
  \hline
  \multirow{2}{*}{Model} & \multicolumn{2}{c|}{Results with 45 training levels} \\
  \cline{2-3}
  & Playable levels & Duplicated levels  \\
  \hline
  Baseline GAN & 19.4\% & 39.4\%\\
  \hline
  CESAGAN & 58\% & 37.6\%\\
  \hline
  \end{tabular}
  \caption{Ratios of playable and duplicated levels with a training set of 45 levels.}
  \label{Tab:Results}
\end{table}

\begin{figure*}[h]
    \centering
    \begin{subfigure}[t]{0.48\linewidth}
        \centering
        \includegraphics[width=0.48\linewidth]{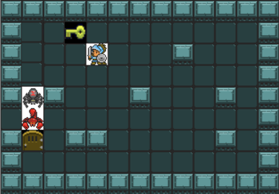}
        \includegraphics[width=0.48\linewidth]{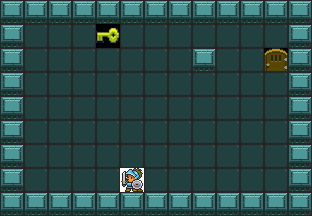}
        \includegraphics[width=0.48\linewidth]{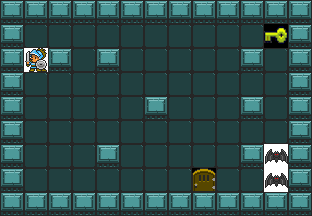}
        \includegraphics[width=0.48\linewidth]{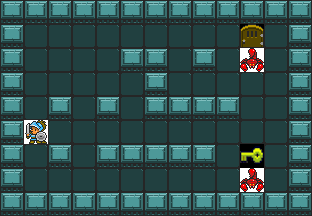}
        \caption{Playable Levels}
        \label{fig:gen_good_levels}
    \end{subfigure}
    \begin{subfigure}[t]{0.48\linewidth}
        \centering
        \includegraphics[width=0.48\linewidth]{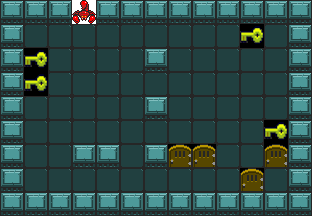}
        \includegraphics[width=0.48\linewidth]{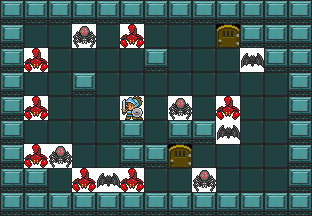}
        \includegraphics[width=0.48\linewidth]{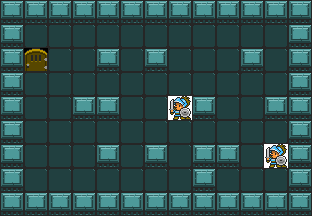}
        \includegraphics[width=0.48\linewidth]{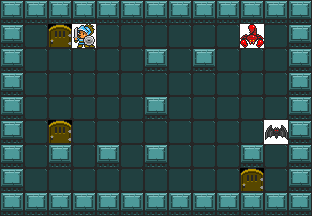}
        \caption{Unplayable Levels}
        \label{fig:gen_bad_levels}
    \end{subfigure}
    \caption{Example of generated levels using CESAGAN with bootstrapping.}
    \label{fig:generated_levels}
\end{figure*}
Table~\ref{Tab:Results} shows the results for CESAGAN without any bootstrapping using a total of 45 Zelda levels as training data. The CESAGANs architecture improves upon both metrics against state-of-art techniques. These results show the potential of the new architecture to improve the playability of generated artifacts. However, the number of duplicate levels has the same order of magnitude with respect to the state-of-art.

\begin{table}
  \centering
  \setlength\tabcolsep{2.2pt}
  \resizebox{\linewidth}{!}{
  \begin{tabular}{|l|c|c|c|}
  \hline
  \multirow{2}{*}{Model} & \multicolumn{2}{c|}{Results with 5 training levels} \\
  \cline{2-3}
   & Playable levels & Duplicated levels  \\
  \hline
  Baseline GAN & 24.6\% & 98\%\\
  \hline
  CESAGAN + boostrapping   & 47\% & 60.3\%\\
  \hline
  \end{tabular}
  }
  \caption{Ratios of playable and duplicated levels using five levels in the training set.}
  \label{Tab:Results 4 levels}
\end{table}

Training on 45 levels is likely not realistic for the majority of video games; just a few human levels are usually available to train for most games. For this reason, CESAGAN with bootstrapping is trained on just five human-designed levels. The results (Table~\ref{Tab:Results 4 levels}) demonstrate that the approach increases the percentage of playable levels, while reducing the number of duplicates considerably.

Figure~\ref{fig:generated_levels} shows eight levels generated using a CESAGAN with bootstrapping and figure~\ref{fig:gen_good_levels} four different playable levels. Besides they satisfy all the constraints, we can notice that they also have small amount of enemies similar to the human designed levels even though the constraint only restricted enemy cover to less than 60\% of the empty tiles. Figure~\ref{fig:gen_bad_levels} displays four unplayable levels that do not follow different constraints but the most notable broken constraints are the numerical constraints (number of avatar, doors, keys, and enemies).

\begin{figure}
    \centering
    \includegraphics[width=0.9\linewidth]{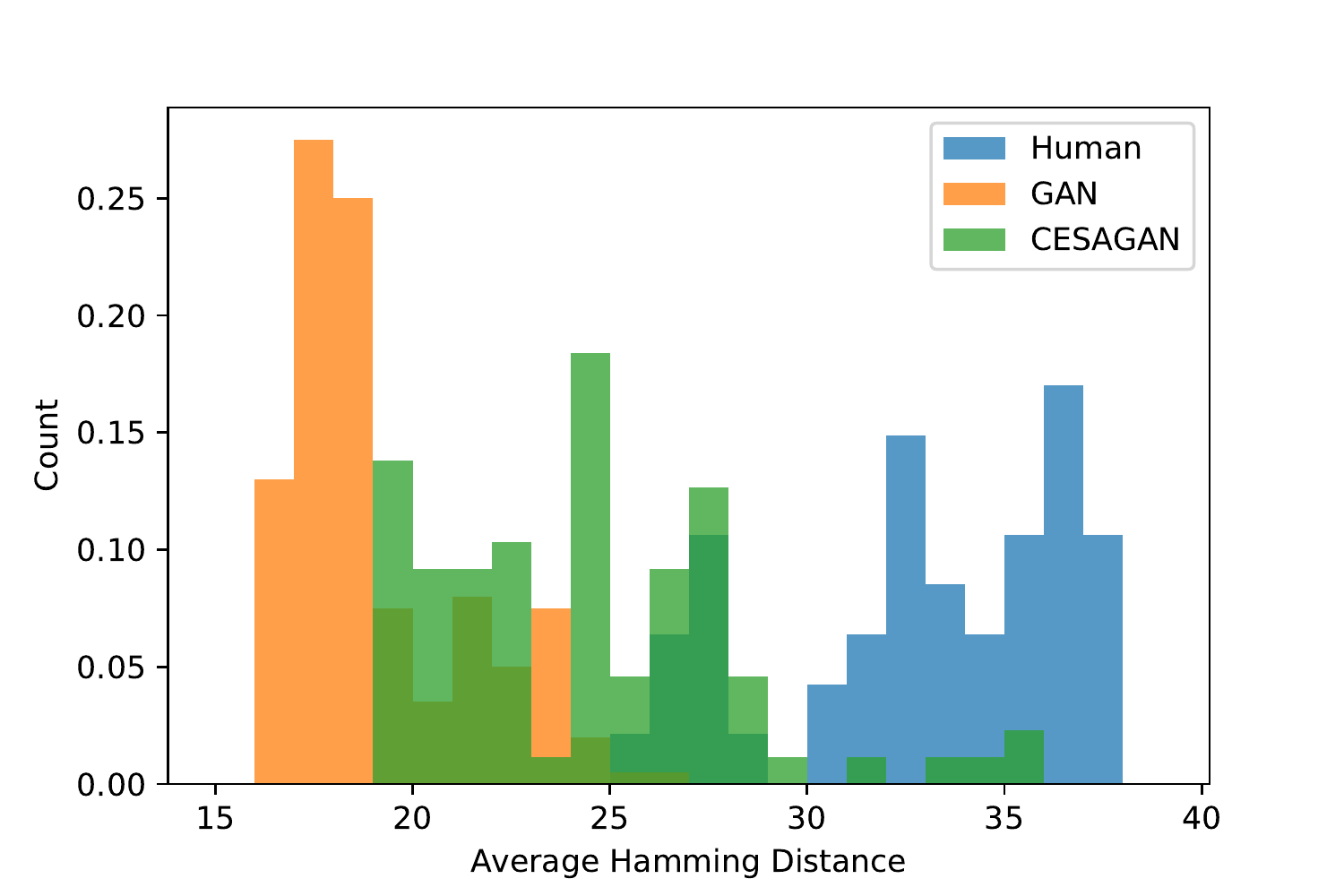}
    \caption{The distribution of the average hamming distance between levels from the same set.}
    \label{fig:hammingDist}
\end{figure}

For further analysis of the generated content, we calculate the average hamming distance between the playable levels for each of the different techniques (GAN and CESAGAN) and compare against the 45 human designed levels. Each level is compared to all the other levels in the same set and the average hamming distance is calculated (number of different tiles). Figure~\ref{fig:hammingDist} shows the average hamming distance between levels from the same set. Human designed levels have the highest hamming distance (mean of $33.31$ and standard deviation of $3.87$), followed by CESAGAN genereated levels (mean of $24.34$ and standard deviation of $3.7$), then GAN generated levels (mean of $19.12$ and standard deviation of $2.3$). A main result is that CESAGAN produces a more diverse set of levels that are as different from each other as possible compared to traditional GANs.

\begin{figure*}[t]
    \centering
    \begin{subfigure}[t]{0.32\linewidth}
        \centering
        \includegraphics[width=1.0\linewidth]{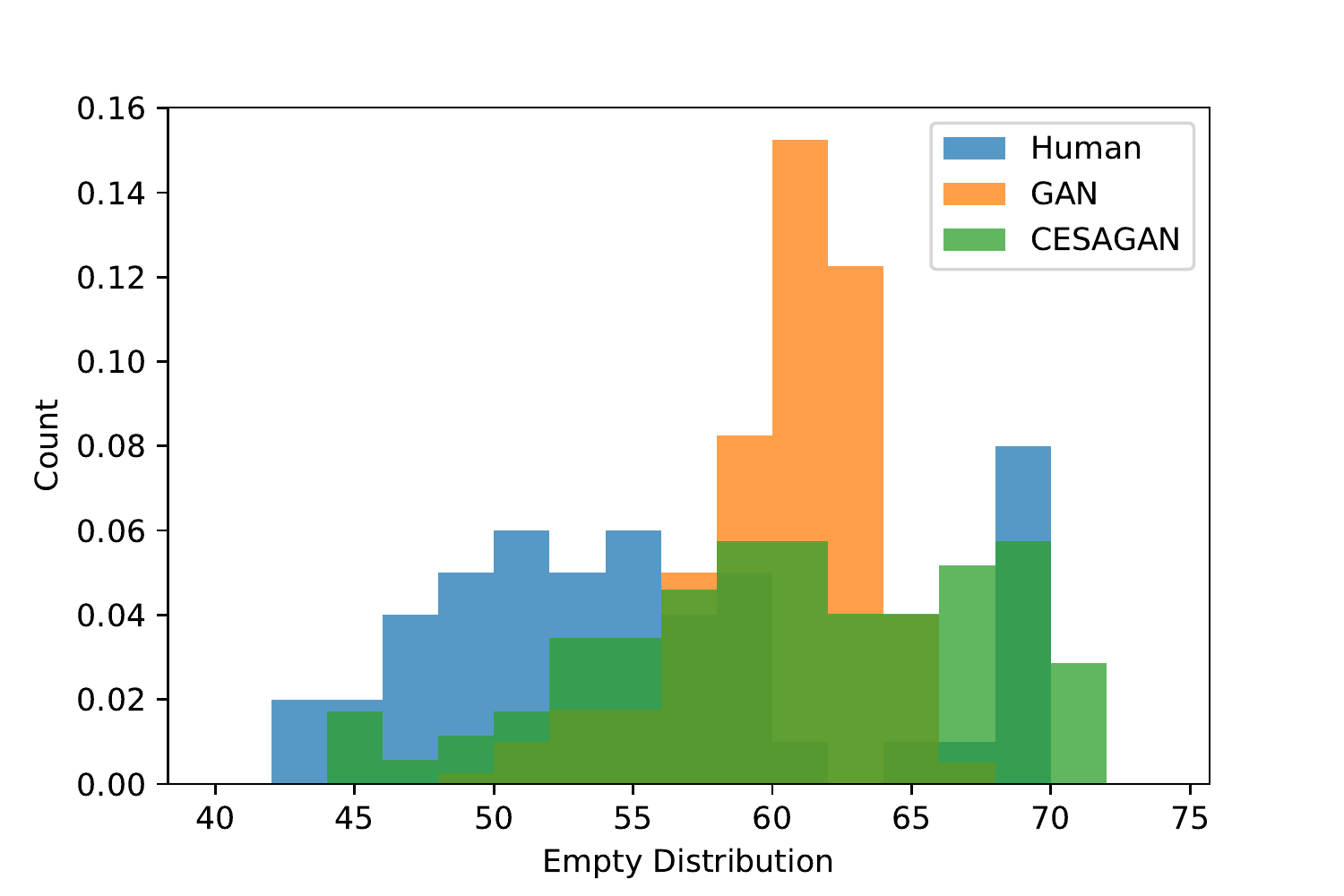}
        \caption{Empty tiles distribution}
        \label{fig:empty_dist}
    \end{subfigure}
    \begin{subfigure}[t]{0.32\linewidth}
        \centering
        \includegraphics[width=1.0\linewidth]{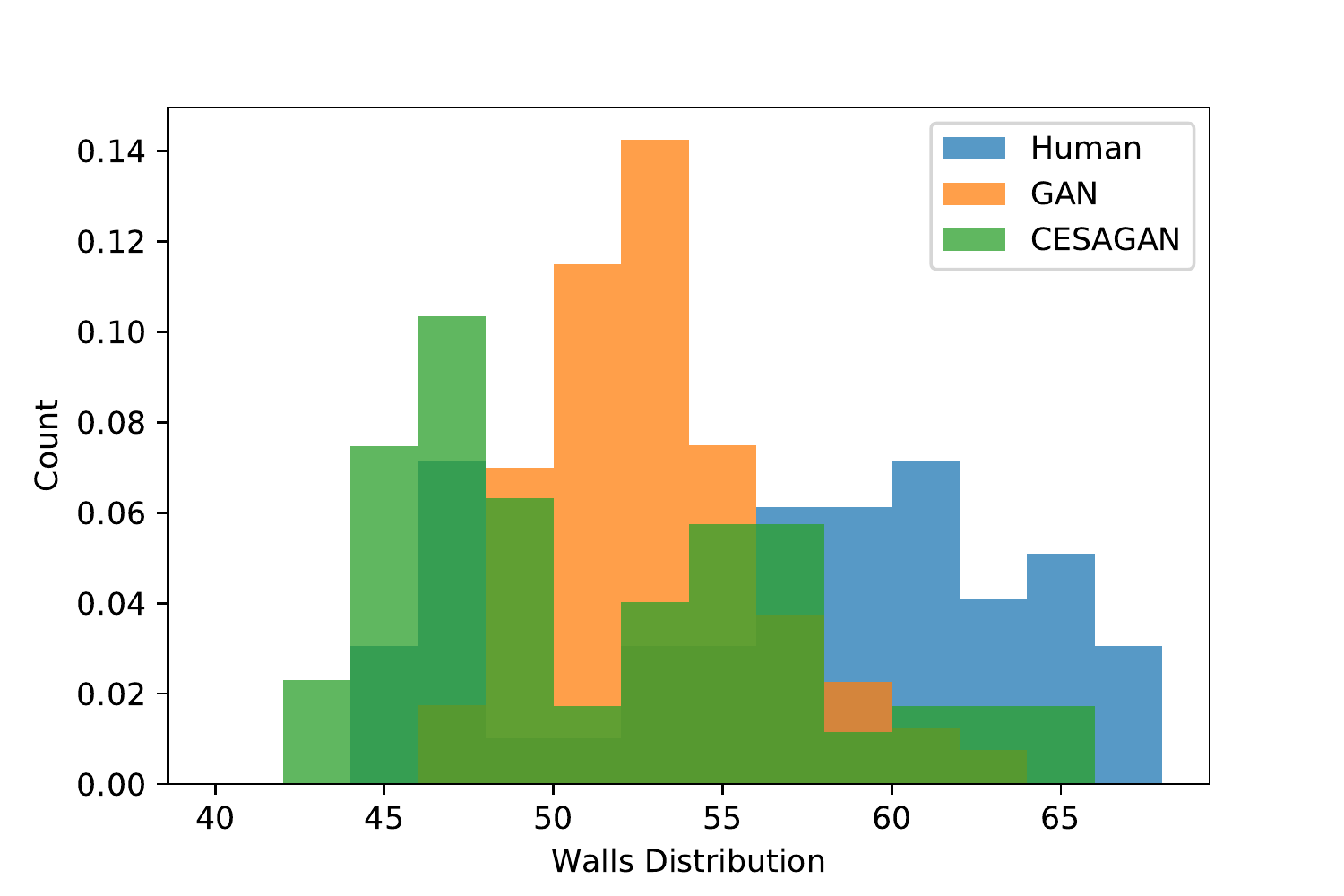}
        \caption{Wall tiles distribution}
        \label{fig:wall_dist}
    \end{subfigure}
    \begin{subfigure}[t]{0.32\linewidth}
        \centering
        \includegraphics[width=1.0\linewidth]{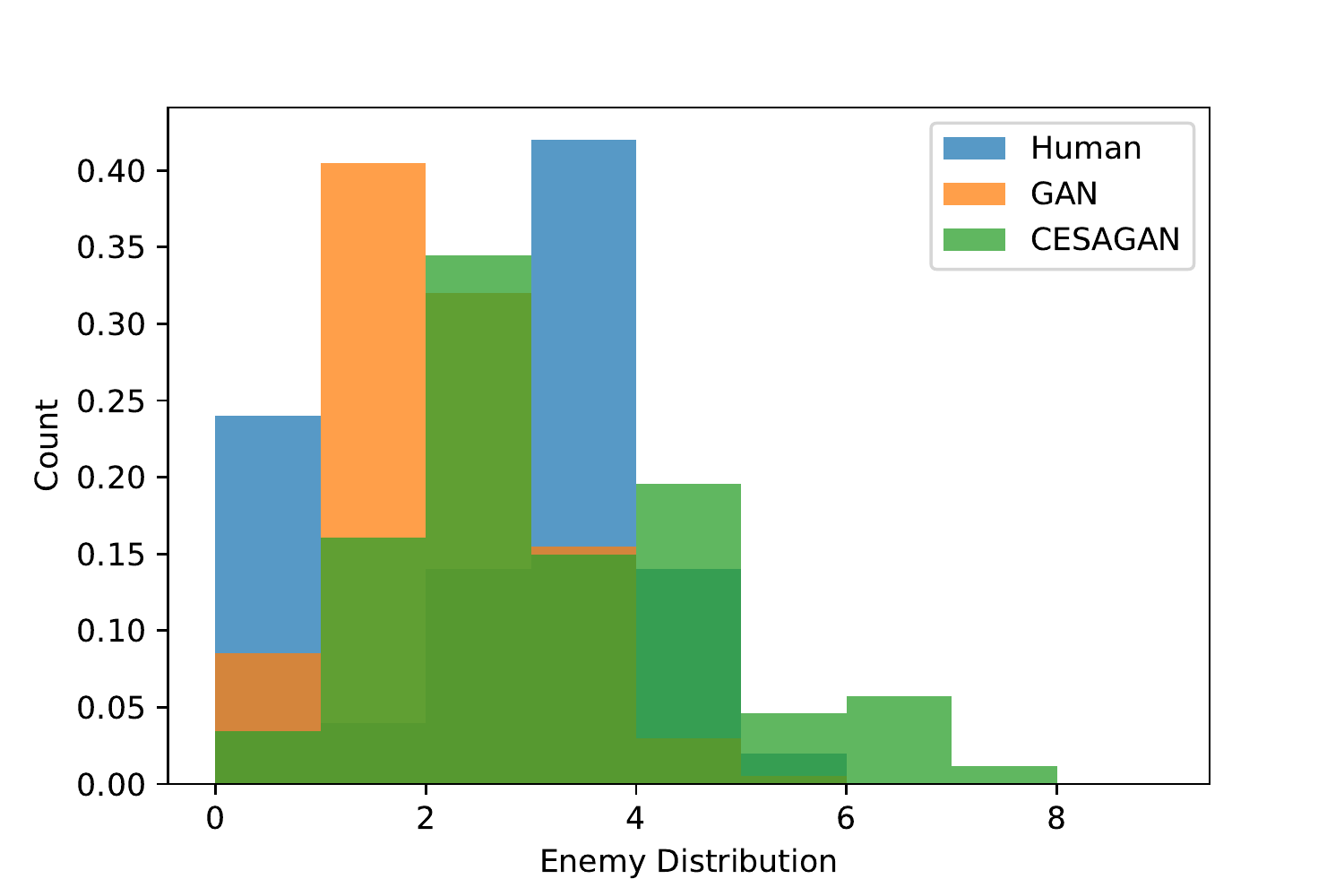}
        \caption{Enemy tiles distribution}
        \label{fig:enemy_dist}
    \end{subfigure}
    \caption{Distribution of different game tiles for CESAGAN, GAN, and Human Levels.}
    \label{fig:dist}
\end{figure*}

We also calculate the distributions of number of different tiles in the playable levels generated CESAGAN and GAN and compare it to the distributions in the human levels. Figure~\ref{fig:dist} shows the different distributions of empty, wall, and enemy tiles in order. We do not show Avatar, Key, or Door tile distribution because they are equal to $1$ in playable levels based on the defined constrained. 
CESAGAN levels have a higher standard deviation compared to GAN levels (nearly double), meaning the CESAGAN model has the ability to generate more diverse levels than the normal GAN.

\section{Conclusion}

We introduce a new GAN architecture --  Conditional Embedding Self-Attention Generative Adversarial Network (CESAGAN) with bootstrapping mechanism -- for video game level generation. 
The results of the experiments confirm the original concern that the state-of-art in GAN has limitations when applied to procedural content generation (PCG). In particular, GANs have difficulty in generating playable and unique levels when few training samples are available. To address this challenge, we introduce Conditional Embedding Self-Attention Generative Adversarial Network (CESAGAN) with bootstrapping. This new architecture is a modification of SAGAN, with an additional feature conditional vector to train the discriminator and generator. The results show a considerable improvement in playability and diversity for 15,000 generated levels with respect to the state-of-art. One of the next challenges for CESAGAN with bootstrapping is to train on more complex video games such as Boulderdash or train with more complex architectures in place of the conditional feature. In addition, using a deep neural network to select the most relevant levels for bootstrapping could decrease the number of duplicate levels even further. 
\label{conclusion}

\section*{Acknowledgements}
Ahmed Khalifa acknowledges the financial support from NSF grant (Award number 1717324 - ``RI: Small: General Intelligence through Algorithm Invention and Selection.''). Michael Cerny Green acknowledges the financial support of the GAANN program. All authors acknowledge Per Josefsen and Nicola Zaltron, who were responsible for the 45 human-designed levels. 
\bibliography{references}

\begin{thebibliography}{}

\bibitem[\protect\citeauthoryear{{Bellemare} \bgroup et al\mbox.\egroup
  }{2013}]{bellemare13arcade}
{Bellemare}, M.~G.; {Naddaf}, Y.; {Veness}, J.; and {Bowling}, M.
\newblock 2013.
\newblock The arcade learning environment: An evaluation platform for general
  agents.
\newblock {\em Journal of Artificial Intelligence Research} 47:253--279.

\bibitem[\protect\citeauthoryear{Bontrager \bgroup et al\mbox.\egroup
  }{2019}]{bontrager2019deepmasterprints}
Bontrager, P.; Roy, A.; Togelius, J.; Memon, N.; and Ross, A.
\newblock 2019.
\newblock Deepmasterprints: Generating masterprints for dictionary attacks via
  latent variable evolution.
\newblock In {\em 2018 IEEE 9th International Conference on Biometrics Theory,
  Applications and Systems (BTAS)},  1--9.
\newblock IEEE.

\bibitem[\protect\citeauthoryear{Bontrager, Togelius, and
  Memon}{2017}]{bontrager2017deepmasterprint}
Bontrager, P.; Togelius, J.; and Memon, N.
\newblock 2017.
\newblock Deepmasterprint: Generating fingerprints for presentation attacks.
\newblock {\em arXiv preprint arXiv:1705.07386}.

\bibitem[\protect\citeauthoryear{Cheng, Dong, and Lapata}{2016}]{cheng2016long}
Cheng, J.; Dong, L.; and Lapata, M.
\newblock 2016.
\newblock Long short-term memory-networks for machine reading.
\newblock {\em arXiv preprint arXiv:1601.06733}.

\bibitem[\protect\citeauthoryear{Dahlskog and
  Togelius}{2012}]{dahlskog2012patterns}
Dahlskog, S., and Togelius, J.
\newblock 2012.
\newblock Patterns and procedural content generation: revisiting mario in world
  1 level 1.
\newblock In {\em Proceedings of the First Workshop on Design Patterns in
  Games}, ~1.
\newblock ACM.

\bibitem[\protect\citeauthoryear{Dahlskog, Togelius, and
  Nelson}{2014}]{dahlskog2014linear}
Dahlskog, S.; Togelius, J.; and Nelson, M.~J.
\newblock 2014.
\newblock Linear levels through n-grams.
\newblock In {\em Proceedings of the 18th International Academic MindTrek
  Conference: Media Business, Management, Content \& Services},  200--206.
\newblock ACM.

\bibitem[\protect\citeauthoryear{Ebner \bgroup et al\mbox.\egroup
  }{2013}]{ebner2013towards}
Ebner, M.; Levine, J.; Lucas, S.~M.; Schaul, T.; Thompson, T.; and Togelius, J.
\newblock 2013.
\newblock Towards a video game description language.
\newblock {\em Dagstuhl Reports}.

\bibitem[\protect\citeauthoryear{Eck and Schmidhuber}{2002}]{eck2002finding}
Eck, D., and Schmidhuber, J.
\newblock 2002.
\newblock Finding temporal structure in music: Blues improvisation with lstm
  recurrent networks.
\newblock In {\em Proceedings of the 12th IEEE workshop on neural networks for
  signal processing},  747--756.
\newblock IEEE.

\bibitem[\protect\citeauthoryear{Ferreira and
  Toledo}{2014}]{ferreira2014search}
Ferreira, L., and Toledo, C.
\newblock 2014.
\newblock A search-based approach for generating angry birds levels.
\newblock In {\em 2014 IEEE Conference on Computational Intelligence and
  Games},  1--8.
\newblock IEEE.

\bibitem[\protect\citeauthoryear{Gaina \bgroup et al\mbox.\egroup
  }{2017}]{gaina2017gvgai}
Gaina, R.~D.; Couetoux, A.; Soemers, D. J. N.~J.; Winands, M. H.~M.; Vodopivec,
  T.; Kirchge$\beta$ner, F.; Liu, J.; Lucas, S.~M.; and Perez-Liebana, D.
\newblock 2017.
\newblock The 2016 two-player {GVGAI} competition.
\newblock {\em IEEE Transactions on Computational Intelligence and AI in
  Games}.

\bibitem[\protect\citeauthoryear{Giacomello, Lanzi, and
  Loiacono}{2018}]{giacomello2018doom}
Giacomello, E.; Lanzi, P.~L.; and Loiacono, D.
\newblock 2018.
\newblock Doom level generation using generative adversarial networks.
\newblock In {\em 2018 IEEE Games, Entertainment, Media Conference (GEM)},
  316--323.
\newblock IEEE.

\bibitem[\protect\citeauthoryear{Goodfellow \bgroup et al\mbox.\egroup
  }{2014}]{goodfellow2014generative}
Goodfellow, I.; Pouget-Abadie, J.; Mirza, M.; Xu, B.; Warde-Farley, D.; Ozair,
  S.; Courville, A.; and Bengio, Y.
\newblock 2014.
\newblock Generative adversarial nets.
\newblock In {\em Advances in neural information processing systems}.

\bibitem[\protect\citeauthoryear{Green \bgroup et al\mbox.\egroup
  }{2018}]{green2018atdelfi}
Green, M.~C.; Khalifa, A.; Barros, G.~A.; Machado, T.; Nealen, A.; and
  Togelius, J.
\newblock 2018.
\newblock Atdelfi: automatically designing legible, full instructions for
  games.
\newblock In {\em Proceedings of the 13th International Conference on the
  Foundations of Digital Games}, ~17.
\newblock ACM.

\bibitem[\protect\citeauthoryear{Guo and Berkhahn}{2016}]{guo2016entity}
Guo, C., and Berkhahn, F.
\newblock 2016.
\newblock Entity embeddings of categorical variables.
\newblock {\em arXiv preprint arXiv:1604.06737}.

\bibitem[\protect\citeauthoryear{Holmgard \bgroup et al\mbox.\egroup
  }{2018}]{holmgard2018automated}
Holmgard, C.; Green, M.~C.; Liapis, A.; and Togelius, J.
\newblock 2018.
\newblock Automated playtesting with procedural personas with evolved
  heuristics.
\newblock {\em IEEE Transactions on Games}.

\bibitem[\protect\citeauthoryear{Hunt, Pong, and
  Tucker}{2007}]{hunt2007difficulty}
Hunt, M.; Pong, C.; and Tucker, G.
\newblock 2007.
\newblock Difficulty-driven sudoku puzzle generation.
\newblock {\em UMAPJournal}  343.

\bibitem[\protect\citeauthoryear{Inc}{2000}]{IDV2000SpeedTree}
Inc, I. D.~V.
\newblock 2000.
\newblock Speedtree.
\newblock \url{https://store.speedtree.com/}.

\bibitem[\protect\citeauthoryear{Karras \bgroup et al\mbox.\egroup
  }{2017}]{karras2017progressive}
Karras, T.; Aila, T.; Laine, S.; and Lehtinen, J.
\newblock 2017.
\newblock Progressive growing of gans for improved quality, stability, and
  variation.
\newblock {\em arXiv preprint arXiv:1710.10196}.

\bibitem[\protect\citeauthoryear{Karth}{2015}]{Karth2015Elite}
Karth, I.
\newblock 2015.
\newblock Elite (1984).
\newblock
  \url{https://procedural-generation.tumblr.com/post/112509130817/elite-1984-elite-created-by-ian-bell-and-david}.

\bibitem[\protect\citeauthoryear{Khalifa and Fayek}{2015}]{khalifa2015puzzle}
Khalifa, A., and Fayek, M.
\newblock 2015.
\newblock Automatic puzzle level generation: A general approach using a
  description language.
\newblock In {\em Computational Creativity and Games Workshop}.

\bibitem[\protect\citeauthoryear{Khalifa \bgroup et al\mbox.\egroup
  }{2016}]{khalifa2016general}
Khalifa, A.; Perez-Liebana, D.; Lucas, S.~M.; and Togelius, J.
\newblock 2016.
\newblock General video game level generation.
\newblock In {\em Proceedings of the Genetic and Evolutionary Computation
  Conference 2016},  253--259.
\newblock ACM.

\bibitem[\protect\citeauthoryear{Khalifa \bgroup et al\mbox.\egroup
  }{2017}]{khalifa2017rulegen}
Khalifa, A.; Green, M.~C.; P{\'e}rez-Li{\'e}bana, D.; and Togelius, J.
\newblock 2017.
\newblock {General Video Game Rule Generation}.
\newblock In {\em 2017 IEEE Conference on Computational Intelligence and Games
  (CIG)}.
\newblock IEEE.

\bibitem[\protect\citeauthoryear{Kodali \bgroup et al\mbox.\egroup
  }{2017}]{kodali2017convergence}
Kodali, N.; Abernethy, J.; Hays, J.; and Kira, Z.
\newblock 2017.
\newblock On convergence and stability of gans.
\newblock {\em arXiv preprint arXiv:1705.07215}.

\bibitem[\protect\citeauthoryear{Lee \bgroup et al\mbox.\egroup
  }{2016}]{lee2016predicting}
Lee, S.; Isaksen, A.; Holmg{\aa}rd, C.; and Togelius, J.
\newblock 2016.
\newblock Predicting resource locations in game maps using deep convolutional
  neural networks.
\newblock In {\em Twelfth Artificial Intelligence and Interactive Digital
  Entertainment Conference}.

\bibitem[\protect\citeauthoryear{Lim and Ye}{2017}]{lim2017geometric}
Lim, J.~H., and Ye, J.~C.
\newblock 2017.
\newblock Geometric gan.
\newblock {\em arXiv preprint arXiv:1705.02894}.

\bibitem[\protect\citeauthoryear{Mirza and
  Osindero}{2014}]{mirza2014conditional}
Mirza, M., and Osindero, S.
\newblock 2014.
\newblock Conditional generative adversarial nets.
\newblock {\em arXiv preprint arXiv:1411.1784}.

\bibitem[\protect\citeauthoryear{Parmar \bgroup et al\mbox.\egroup
  }{2018}]{parmar2018image}
Parmar, N.; Vaswani, A.; Uszkoreit, J.; Kaiser, {\L}.; Shazeer, N.; Ku, A.; and
  Tran, D.
\newblock 2018.
\newblock Image transformer.
\newblock {\em arXiv preprint arXiv:1802.05751}.

\bibitem[\protect\citeauthoryear{Perez-Liebana \bgroup et al\mbox.\egroup
  }{2016a}]{perez2016general}
Perez-Liebana, D.; Samothrakis, S.; Togelius, J.; Lucas, S.~M.; and Schaul, T.
\newblock 2016a.
\newblock {General Video Game AI: Competition, Challenges and Opportunities}.
\newblock In {\em Thirtieth AAAI Conference on Artificial Intelligence},
  4335--4337.

\bibitem[\protect\citeauthoryear{Perez-Liebana \bgroup et al\mbox.\egroup
  }{2016b}]{perez20162014}
Perez-Liebana, D.; Samothrakis, S.; Togelius, J.; Schaul, T.; Lucas, S.~M.;
  Cou{\"e}toux, A.; Lee, J.; Lim, C.-U.; and Thompson, T.
\newblock 2016b.
\newblock The 2014 general video game playing competition.
\newblock {\em IEEE Transactions on Computational Intelligence and {AI} in
  Games} 8(3):229--243.

\bibitem[\protect\citeauthoryear{Perez-Liebana \bgroup et al\mbox.\egroup
  }{2018}]{perez2018general}
Perez-Liebana, D.; Liu, J.; Khalifa, A.; Gaina, R.~D.; Togelius, J.; and Lucas,
  S.~M.
\newblock 2018.
\newblock General video game {AI}: a multi-track framework for evaluating
  agents, games and content generation algorithms.
\newblock {\em arXiv preprint arXiv:1802.10363}.

\bibitem[\protect\citeauthoryear{Radford \bgroup et al\mbox.\egroup
  }{2018}]{radford2018better}
Radford, A.; Wu, J.; Amodei, D.; Amodei, D.; Clark, J.; Brundage, M.; and
  Sutskever, I.
\newblock 2018.
\newblock Better language models and their implications.

\bibitem[\protect\citeauthoryear{Salimans \bgroup et al\mbox.\egroup
  }{2016}]{salimans2016improved}
Salimans, T.; Goodfellow, I.; Zaremba, W.; Cheung, V.; Radford, A.; and Chen,
  X.
\newblock 2016.
\newblock Improved techniques for training gans.
\newblock In {\em Advances in neural information processing systems},
  2234--2242.

\bibitem[\protect\citeauthoryear{Shaker, Togelius, and
  Nelson}{2016}]{shaker2016procedural}
Shaker, N.; Togelius, J.; and Nelson, M.~J.
\newblock 2016.
\newblock {\em Procedural content generation in games}.
\newblock Springer.

\bibitem[\protect\citeauthoryear{Snodgrass and
  Ontan{\'o}n}{2016}]{snodgrass2016controllable}
Snodgrass, S., and Ontan{\'o}n, S.
\newblock 2016.
\newblock Controllable procedural content generation via constrained
  multi-dimensional markov chain sampling.
\newblock In {\em IJCAI},  780--786.

\bibitem[\protect\citeauthoryear{Summerville and
  Mateas}{2016a}]{summerville2016super}
Summerville, A., and Mateas, M.
\newblock 2016a.
\newblock Super mario as a string: Platformer level generation via lstms.
\newblock In {\em Foundations of Digital Games}.

\bibitem[\protect\citeauthoryear{Summerville and
  Mateas}{2016b}]{summerville2016mystical}
Summerville, A.~J., and Mateas, M.
\newblock 2016b.
\newblock Mystical tutor: A magic: The gathering design assistant via denoising
  sequence-to-sequence learning.
\newblock In {\em Twelfth Artificial Intelligence and Interactive Digital
  Entertainment Conference}.

\bibitem[\protect\citeauthoryear{Summerville \bgroup et al\mbox.\egroup
  }{2018}]{summerville2018procedural}
Summerville, A.; Snodgrass, S.; Guzdial, M.; Holmg{\aa}rd, C.; Hoover, A.~K.;
  Isaksen, A.; Nealen, A.; and Togelius, J.
\newblock 2018.
\newblock Procedural content generation via machine learning (pcgml).
\newblock {\em IEEE Transactions on Games} 10(3):257--270.

\bibitem[\protect\citeauthoryear{Taylor and
  Parberry}{2011}]{taylor2011procedural}
Taylor, J., and Parberry, I.
\newblock 2011.
\newblock Procedural generation of sokoban levels.
\newblock In {\em Proceedings of the International North American Conference on
  Intelligent Games and Simulation},  5--12.

\bibitem[\protect\citeauthoryear{Togelius \bgroup et al\mbox.\egroup
  }{2011}]{togelius2011search}
Togelius, J.; Yannakakis, G.~N.; Stanley, K.~O.; and Browne, C.
\newblock 2011.
\newblock Search-based procedural content generation: A taxonomy and survey.
\newblock {\em IEEE Transactions on Computational Intelligence and AI in Games}
  3(3):172--186.

\bibitem[\protect\citeauthoryear{Vaswani \bgroup et al\mbox.\egroup
  }{2017}]{vaswani2017attention}
Vaswani, A.; Shazeer, N.; Parmar, N.; Uszkoreit, J.; Jones, L.; Gomez, A.~N.;
  Kaiser, {\L}.; and Polosukhin, I.
\newblock 2017.
\newblock Attention is all you need.
\newblock In {\em Advances in neural information processing systems},
  5998--6008.

\bibitem[\protect\citeauthoryear{Volz \bgroup et al\mbox.\egroup
  }{2018}]{volz2018evolving}
Volz, V.; Schrum, J.; Liu, J.; Lucas, S.~M.; Smith, A.; and Risi, S.
\newblock 2018.
\newblock Evolving mario levels in the latent space of a deep convolutional
  generative adversarial network.
\newblock In {\em Proceedings of the Genetic and Evolutionary Computation
  Conference},  221--228.
\newblock ACM.

\bibitem[\protect\citeauthoryear{Zhang \bgroup et al\mbox.\egroup
  }{2018}]{zhang2018self}
Zhang, H.; Goodfellow, I.; Metaxas, D.; and Odena, A.
\newblock 2018.
\newblock Self-attention generative adversarial networks.
\newblock {\em arXiv preprint arXiv:1805.08318}.

\end{thebibliography}
\bibliographystyle{aaai}

\end{document}